\documentclass[12pt,journal,transmag,onecolumn]{IEEEtran}
\usepackage{graphicx}
\usepackage{subfigure}
\usepackage{amsmath,amssymb} 
\usepackage{algorithm}  
\usepackage{algorithmic} 
\usepackage{fancyhdr}
\usepackage{multirow}
\usepackage{multicol}
\usepackage{color}
\usepackage{marvosym}
\usepackage{graphics,graphicx}
\usepackage{float}
\usepackage{color}
\usepackage{cite}

\begin{document}
	
	\title{ An optimal hierarchical clustering approach to segmentation of mobile LiDAR point clouds}

	\author{\IEEEauthorblockN{Sheng Xu, Ruisheng Wang, Han Zheng \IEEEauthorrefmark{}}
		\IEEEauthorrefmark{}Department of Geomatics Engineering, University of Calgary, Alberta, Canada}
	
	\IEEEtitleabstractindextext{%
		\begin{abstract}
					
			This paper proposes a hierarchical clustering approach for the segmentation of mobile LiDAR point clouds. We perform the hierarchical clustering on unorganized point clouds based on a proximity matrix. The dissimilarity measure in the proximity matrix is calculated by the Euclidean distances between clusters and the difference of normal vectors at given points. The main contribution of this paper is that we succeed to optimize the combination of clusters in the hierarchical clustering. The combination is obtained by achieving the matching of a bipartite graph, and optimized by solving the minimum-cost perfect matching. Results show that the proposed optimal hierarchical clustering (OHC) succeeds to achieve the segmentation of multiple individual objects automatically and outperforms the state-of-the-art LiDAR point cloud segmentation approaches.
			
		\end{abstract}
	}
	
	\maketitle
	\thispagestyle{fancy}
	\renewcommand{\headrulewidth}{0pt}
	\fancyhead[L]{}
	\fancyhead[R]{}
	\fancyfoot[C]{\thepage}
	\pagestyle{plain}
	
	\section{Introduction}
		
	Segmentation is the process of partitioning input data into multiple individual objects. In general, an individual object is defined as a region with uniform attribute. In point cloud segmentation, the results can be divided into three levels, namely the supervoxel level\cite{wang20153}, primitive level (e.g. line\cite{xia2017fast}, plane \cite{nan2017polyfit} and cylinder \cite{qiu2014pipe}), and object level \cite{YuLiGuanEtAl2015pp13741386,douillard2011segmentation,GolovinskiyKimFunkhouser2009pp21542161,KlasingWollherrBuss2008pp40434048}. 
	
	Due to the fact that point clouds are noisy, uneven and unorganized, the accuracy of the segmentation is far from being desired, especially in the segmentation of multiple overlapping objects. This paper aims to propose a hierarchical clustering approach for the segmentation of mobile LiDAR point clouds in the object level. In order to obtain the optimal combination of clusters, we use the matching in a graph to indicate the combination and achieve the optimal solution by solving the minimum-cost perfect matching in a bipartite graph. 
	 	
	This paper is organized as follows. Section \uppercase\expandafter{\romannumeral2} reviews merits and demerits of the related segmentation methods. Section \uppercase\expandafter{\romannumeral3} overviews the procedure of the proposed optimal hierarchical clustering approach. Section \uppercase\expandafter{\romannumeral4}  calculates the proximity matrix to store the dissimilarity of clusters. Section \uppercase\expandafter{\romannumeral5} presents details of the clustering optimization. Section \uppercase\expandafter{\romannumeral6} shows experiments to evaluate the performance of the proposed algorithm. The conclusions are outlined in Section \uppercase\expandafter{\romannumeral7}.
	
	\section{Related work}
	
	Segmentation from point clouds collected by laser sensors with an accurate 3D information plays an important role in the environmental analysis, 3D modeling and object tracking. In the following, related methods proposed for the segmentation of point clouds will be reviewed and analyzed.
	
	In the work of Douillard et al. \cite{douillard2011segmentation}, they propose a Cluster-All method for the segmentation of dense point clouds and achieve a trade-off in terms of the simplicity, accuracy and computation time. Their Cluster-All is performed based on the $k$-nearest neighbors approach. The principle of KNNiPC ($k$-nearest neighbors in point clouds) is to select a number of points in the nearest Euclidean distance for a given point and assign them the same group index. To decrease the under-segmentation rate in results, the variance and mean of the Euclidean distances of the points within a group are restricted to be less than a preset threshold value. A similar approach is shown in the work of Klasing et al.\cite{KlasingWollherrBuss2008pp40434048}. They present a radially nearest neighbor strategy to bound the region of neighbor points which helps enhance the robustness of the neighbor selection. KNNiPC works well in different terrain areas and does not require any prior knowledge of the location of objects. The problem is that results of KNNiPC highly depend on the selection of a ``good value'' for $k$. 
	 
	In the segmentation of point clouds, the challenging is to separate two overlapping objects.  
	In the method of Yu et al. \cite{YuLiGuanEtAl2015pp13741386} , they propose an algorithm for segmenting street light poles from mobile point clouds. In order to separate a cluster with more than one object, they add the elevation information to extend the 2D normalized-cut method \cite{ShiMalik2000pp888905}. Their proposed 3DNCut (3D normalized-cut) partitions input points into two disjoint groups by minimizing the similarity within each group and maximizing the dissimilarity between different groups. 3DNCut obtains an optimal solution of the binary segmentation problem which demonstrates a promising method for mobile LiDAR point cloud segmentation. 
	The shortcoming is that the number of objects has to be preset manually in the multi-label segmentation task.  In the method of Golovinskiy et al.\cite{GolovinskiyKimFunkhouser2009pp21542161}, they present a min-cut based method (MinCut) for segmenting objects from point clouds. MinCut partitions input points into two disjoint groups, i.e. background and foreground, by minimizing their energy function. The solution cut, which is used to separate the input scene into background and foreground, is obtained by the graph cut method \cite{BoykovFunka-Lea2006pp109131}. MinCut obtains competitive segmentation results in terms of the optimum and accuracy. However, MinCut requires for the location of each object in the multi-object segmentation task. To achieve the desired performance, users have to set the center point and radius for different objects manually.
		
	Clustering is a well-known technique in data mining. It aims to find borders between groups based on defined measures, such as the density and Euclidean distance. The goal is that data in the same group are more similar to each other than to those in other groups. If we use one called `object distance' as the dissimilarity measure in the clustering, the result of the clustering will coincide with the object segmentation in point clouds. The clustering approach may provide a new perspective in the segmentation of individual objects.
		
	In general, clustering methods are divided into five groups, including partitioning, hierarchical, density-based, grid-based or combinations of these \cite{han2011data}. Both density-based \cite{du2016study,xie2016robust} and grid-based methods \cite{yanchang2001gdilc} highly rely on the assumption that cluster centers have a higher density than their neighbors and each center has a large distance from points with higher densities. Most density- and grid-based algorithms require users to specify density thresholds for choosing cluster centers. However, the density of the LiDAR data depends on the sensor and the beam travel distance. Points within an object may have various densities, thus, density- and grid-based clustering methods are not suitable for achieving the multi-object segmentation from LiDAR point clouds.
	
	The classical partition method is the $k$-means approach, which has been used for the point cloud segmentation in  \cite{lavoue2005new,isprs-archives-XLII-1-W1-595-2017}. The idea of KMiPC ($k$-means in point clouds) is to partition points into different sets to minimize the sum of distances of each point in the cluster to the center. In the work of Lavoue et al. \cite{lavoue2005new}, they use $k$-means for the segmentation of ROI (region of interest) from mesh points. The problem is that KMiPC is easy to produce redundant segments as shown in \cite{lavoue2005new}. Therefore, a merging process to reduce the over-segmentation rate is required. Moreover, KMiPC asks for setting the number of clusters manually and tends to segment groups evenly as shown in \cite{isprs-archives-XLII-1-W1-595-2017}. In the work of Feng et al. \cite{feng2014fast}, they propose a plane extraction method based on the agglomerative clustering approach (PEAC). They segment planes from point clouds efficiently. PEAC can be used for the individual object segmentation by merging clusters belonging to the same object. The shortcoming is that their input point clouds are required to be well-organized.
	
	This paper aims to propose a new bottom-up hierarchical clustering algorithm for the individual object segmentation from LiDAR point clouds. Our agglomerative clustering starts with a number of clusters and each of them contains only one point.  A series of merging operations are then followed to combine similar clusters into one cluster to move up the hierarchy. An individual object will be lead to the same group in the top hierarchy. The proposed hierarchical clustering algorithm does not request for presetting the number of clusters or inputting the location of each object, which is significant in the segmentation of multiple objects. The combination of clusters in the hierarchical clustering is optimized by solving the minimum-cost perfect matching of a bipartite graph. 
		
	\section{Methodology}
	
	This section will show details of our optimal hierarchical clustering (OHC) algorithm for the point cloud segmentation.  First, we show the procedure of the proposed hierarchical clustering to give an overview of our methodology. Then, we focus on the calculation of the dissimilarity to form the proximity matrix for optimizing the cluster combination. Finally, we use the matching in a graph to indicate the combination solution and achieve the optimal combination via solving the minimum-cost perfect matching in a bipartite graph. 		
		
	\subsection{The procedure of the proposed OHC}
	
	Assume that the input point set is $P=\{p_1,p_2,...,p_n\}$ and the cluster set is $C=\{c_1,c_2,...c_i,...,c_j,...,c_t\}$. $n$ is the number of points in $P$ and $t$ is the number of clusters in $C$. Each cluster $c_i$ contains one or more points from $P$, and $c_i \cap c_j$ is $\O$ under $i \ne j$. The goal of OHC is to optimize the set $C$ to achieve that each $c_i \in C$ is a cluster of points of an individual object. A brief overview of our OHC is as follows. 
	
	(1) Start with a cluster set $C$ and each cluster $c_i$ consists of a point $p_i \in P$;
	
	(2) Calculate the proximity matrix by measuring the dissimilarity of clusters in $C$;
	
	(3) Calculate the optimal solution for merging clusters by  
	
	\begin{align}
	\arg\min_{\Omega}\sum_{M_i \in \Omega } D(M_i)
	\label{cluster}
	\end{align}
	
	where $M_i=\{c_a,c_b\}$ means a pair of clusters, and $c_a \in C$ and $c_b \in C$. $\Omega=\{M_1,M_2,...,M_i,...\}$ is the set of the pair of clusters and $M_i \cap  M_j= \O$. $D(*)$ is the dissimilarity of two clusters stored in the proximity matrix;
		
	(4) Combine clusters in each $M_i\in \Omega$ into one cluster and use those combined clusters to update the set $C$;
		
	(5) Repeat (2)-(4) until $C$ converges.
	
	In the initialization, each cluster $c_i$ contains only one point from $P$, $t$ is equal to $n$ and the size of the proximity matrix is $t \times t$. In the hierarchical clustering, similar clusters are combined into one cluster and thus the value of $t$ will be reduced in the clustering process. The key steps in the procedure are (2) and (3) which will be discussed in the following sections. 
	
	\subsection{The calculation of the proximity matrix}
	For a set with $t$ clusters, we define a $t \times t$ symmetric matrix called the proximity matrix.  The $(i,j)$th element of the matrix means the dissimilarity measure for the $i$th and $j$th clusters $(i,j=1,2,...,t)$.  In the hierarchical clustering, two clusters with a low dissimilarity are preferred to be combined into one cluster. 
	
	In our work, the calculation of the dissimilarity contains a distance measure $\alpha \left(p_i,p_j,c_{i},c_{j}\right)$ and a direction measure $\beta\left(p_i,p_j,c_{i},c_{j}\right)$. The distance term $\alpha \left(p_i,p_j,c_{i},c_{j}\right)$ is formed as  
	
	\begin{align}	
	\alpha\left(p_i,p_j,c_{i},c_{j}\right)=\frac{ d(p_i,p_j)}{\max(m(c_i),m(c_j))}
	\label{alpha}
	\end{align}
	where $d(*,*)$ is the Euclidean distances of two points. Clusters $c_i$ and $c_j$ are two different clusters in the set $C$. Points $p_i$ and $p_j$ are obtained by
	
	\begin{align}
	(p_i,p_j)=\arg\min_{(p,p')}{d(p,p'): p \in c_i, p' \in c_j}
	\label{min}
	\end{align}
		
	The density of LiDAR points relies on the scanner sensor and the beam travel distance, therefore, the magnitude of distances of points are various in different groups. In the calculation of $\alpha\left(p_i,p_j,c_{i},c_{j}\right)$, we use $m(*)$, which is based on a median filter \textbf{Med} and defined by Eq.(\ref{median}), to normalized the Euclidean distance values of points in different groups. 
	\begin{align}
	m(c_i)= \textbf{Med}_{p \in c_i}\{ \min_{p' \in c_i, p' \ne p} d(p,p')\}
	\label{median}
	\end{align}
			
	\noindent For example, if we have a cluster $c_0=\{p_1, p_2, p_3\}$ and the minimal distance between the point $p_i \in c_0$ and other points is $d_i$, our $m(c_0)$ is obtained by the median value of $\{d_1,d_2,d_3\}$.
	
	Our direction measure is based on the normal vector information. In our work, the normal vector at a point is approximated with the normal to its $k$-neighborhood surface by performing PCA (Principal Component Analysis) on the neighborhood’s covariance matrix \cite{rusu2010semantic}.  The direction term $\beta\left(p_i,p_j,c_{i},c_{j}\right)$ is formed as

	\begin{align}
	\beta\left(p_i,p_j,c_{i},c_{j}\right)=1-\left | \textbf{V}(p_i) \cdot \textbf{V}(p_j)\right |
	\label{beta} 
	\end{align}
	\noindent where $\textbf{V}(p_i)$ and $\textbf{V}(p_j)$ are normal vectors estimated from the $k$-nearest neighbor points of $p_i$ and $p_j$, respectively. Points $p_i$ and $p_j$ are obtained in Eq.(\ref{min}).
	
	According to our prior knowledge, the object's exterior points are potential to be the overlapping points of different objects and interior points are mostly from the same object. Thus, the proximity matrix $\textbf{PM}$ is defined as
		
	\begin{align}
	\textbf{PM}(c_i,c_j)=\left\{\begin{matrix}
	\frac{\lambda-1}{\lambda} \alpha\left(p_i,p_j,c_{i},c_{j}\right)+\frac{1}{\lambda}  \beta\left(p_i,p_j,c_{i},c_{j}\right),& \delta_{p_i,p_j}=1 \\ 
	\frac{1}{\lambda} \alpha\left(p_i,p_j,c_{i},c_{j}\right)+\frac{\lambda-1}{\lambda}  \beta\left(p_i,p_j,c_{i},c_{j}\right),&\delta_{p_i,p_j}=0 \\
	\frac{1}{2} \alpha\left(p_i,p_j,c_{i},c_{j}\right)+\frac{1}{2} \beta\left(p_i,p_j,c_{i},c_{j}\right),&\delta_{p_i,p_j}=0.5 
	\end{matrix}\right.
	\label{smoothnessterm}
	\end{align}
	
	\noindent where $\lambda$ is a user-defined weight coefficient to balance the distance measure $\alpha\left(p_i,p_j,c_{i},c_{j}\right)$ and the direction measure $\beta\left(p_i,p_j,c_{i},c_{j}\right)$. The $\delta_{p_i,p_j}$ is to predict the region of the point $p_i$ and $p_j$. If both $p_i$ and $p_j$ are interior points, which are assumed to be in the same cluster, $\delta_{p_i,p_j}=1$; if both of them are exterior points, which are assumed to be in different clusters, $\delta_{p_i,p_j}=0$; in all other conditions, $\delta_{p_i,p_j}=0.5$. In our work, the dissimilarity of two clusters is small if they are spatially close or normal vectors at the given points are consistent. When both two clusters consist of interior points, the dissimilarity is dominated by the distance measure. When both two clusters consist of exterior points, the direction measure dominates their dissimilarity.

	To calculate $\delta_{p_i,p_j}$, we propose a local 3D convex hull testing method to mark the exterior and interior points. The proposed testing proceeds as follows: (1) initialize all input points as unlabeled points; (2) check if a point is not labeled as an interior point; (3) pick up its $k$-nearest neighbor points to construct a local 3D convex hull; (4) label all points inside the hull as interior points; (5) repeat (2)-(4) until all input points are tested; (6) mark the rest unlabeled points as exterior points. 
	
	\begin{figure}
		\centering
		\includegraphics[width=16cm]{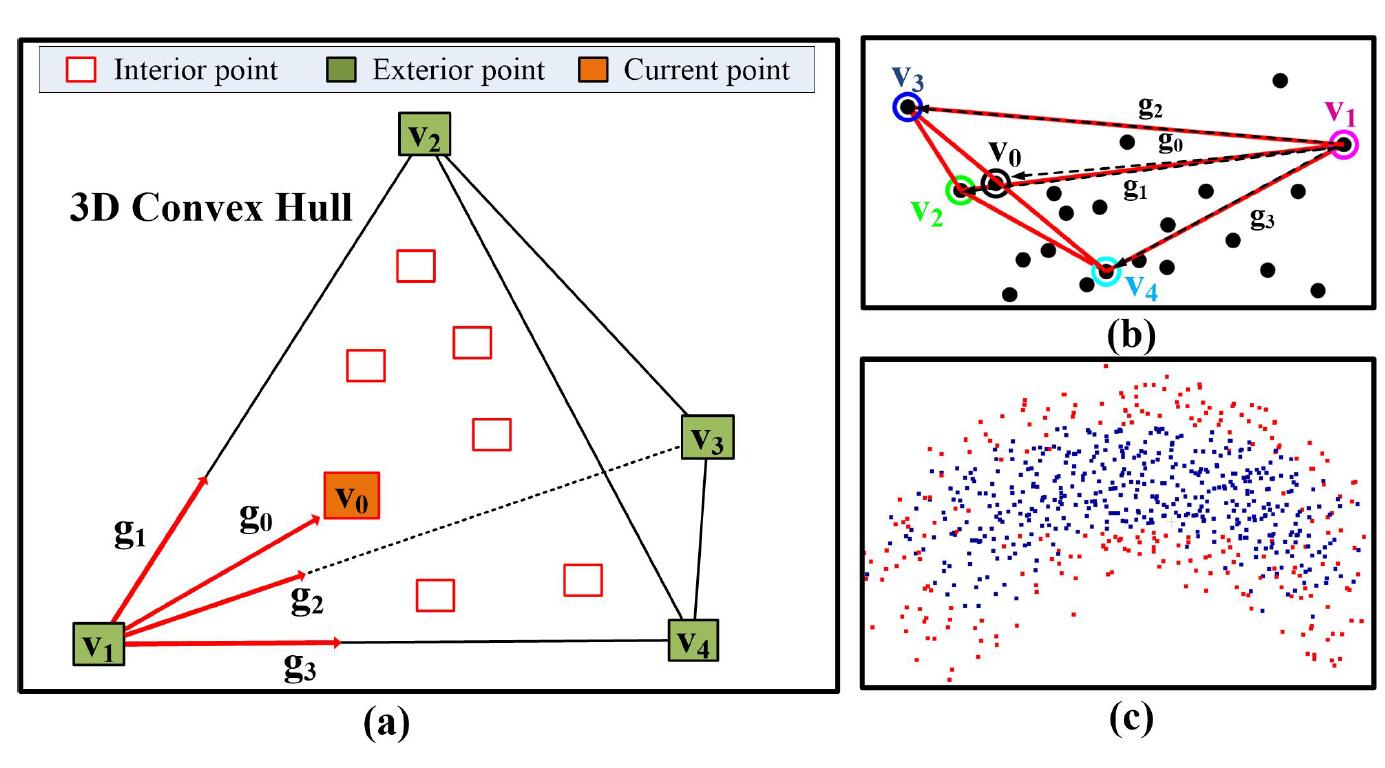} 
		\caption{ Illustration of the local 3D convex hull testing. (a) The testing of a vertex $v_0$. (b)  A close-up view of a local 3D convex hull. (c) Extracted exterior points (red) and interior points (blue).}
		\label{points}
 \end{figure}
	
	To implement the testing, we form a local 3D convex hull using four vertices, namely $v_{1}$, $v_{2}$, $v_{3}$ and $v_{4}$, as shown in Fig.\ref{points}(a). The target is to test if a vertex $v_{0}$ is an interior point or not. As shown in Fig.\ref{points}, the vector $\textbf{g}_0=(v_{0}-v_{1})$ can be represented by the sum of vectors $\textbf{g}_1=(v_2-v_1)$, $\textbf{g}_2=(v_3-v_1)$ and $\textbf{g}_3=(v_4-v_1)$ as
	
	\begin{align}
	\textbf{g}_0=u \times \textbf{g}_1 + v \times \textbf{g}_2 + w \times \textbf{g}_3
	\label{weightcenter}
	\end{align}
	
	\noindent Based on Eq.(\ref{weightcenter}), one can conclude that if $u \nless 0, v\nless 0, w \nless 0$ and $u+v+w<1$, $v_{0}$ will be inside the 3D convex hull. Our testing relies on values of $u$, $v$ and $w$, which can be solved efficiently by
	
	$$\left[u,v,w\right]^\top=\left[\textbf{g}_1,\textbf{g}_2,\textbf{g}_3\right]^{-1}\cdot \textbf{g}_0$$
	
	If $v_{0}$ is inner the 3D convex hull, it will be labeled as an interior point. Each point from the input will be selected as a vertex and tested in a local convex hull constructed by its $k$-nearest neighbor points. In building the 3D convex hull, $v_{1}$ is chosen as the furthest point to $v_{0}$, $v_{2}$ is the point to obtain the largest projection of $\textbf{g}_1$ on the $\textbf{g}_0$ direction (i.e. $\arg\max\limits_{v_2}(|\textbf{g}_1|\cdot\cos<\textbf{g}_1,\textbf{g}_0>)$), $v_{3}$ is the furthest point to the line containing $v_{2}$ and $v_{1}$, and $v_{4}$ is the furthest point to the plane containing $v_{1}$, $v_{2}$ and $v_{3}$. A close-up view of a local 3D convex hull is shown in Fig.\ref{points}(b) and Fig.\ref{points}(c) shows the extracted interior points (blue) and exterior points (red) from a test set. 
	
	\subsection{The optimization of the clustering}
		
	Usually, the hierarchal clustering uses a greedy strategy in the combination of clusters.  The idea is to find the most similar two clusters based on the proximity matrix first, and then combine them into one cluster. Repeat the above steps until all objects are grouped in the same cluster. The greedy strategy is easy to incur a local optimization. This section targets at optimizing the combination of clusters globally.  
	
\begin{figure}
		\centering
		\includegraphics[width=8cm]{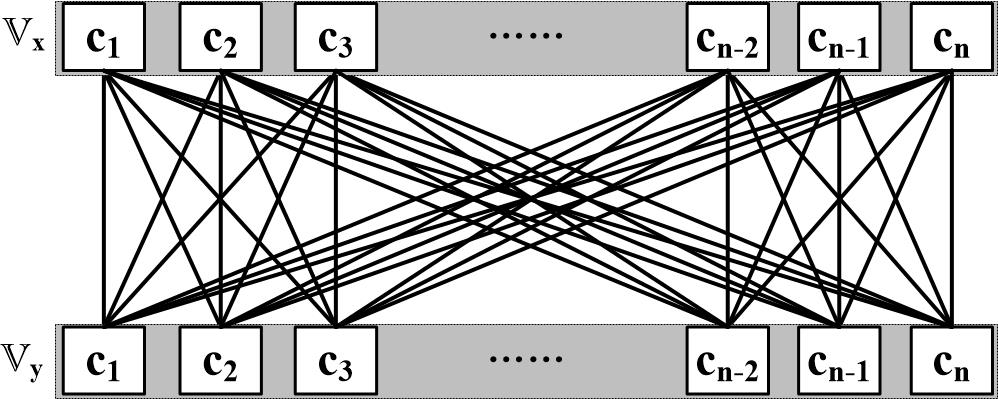}
		\caption{The modeled bipartite graph $\mathbb{G}=\left \{ \mathbb{V}_{x}, \mathbb{V}_{y}, \mathbb{E} \right \}$.}
		\label{rawsegmentation}
	\end{figure}

	Denote our graph as $\mathbb{G}=\left \{ \mathbb{V}_{x}, \mathbb{V}_{y}, \mathbb{E} \right \}$, where the node set $\mathbb{V}_{x}=\{c_{1},c_{2},...,c_{i},...,c_{n}\}$ describes the current clusters in the set $C$ and $\mathbb{V}_{y}$ shows clusters for the combination. In the hierarchical clustering, any two clusters can be combined into one cluster, therefore, we let $\mathbb{V}_{y}=\mathbb{V}_{x}$. The edge set $\mathbb{E}=\{e_{1,1}, e_{1,2},....,e_{1,n},e_{2,1}, ....,e_{2,n},...,e_{i,j},...,e_{n,1}, ..., e_{n,n}\}$ shows connections between the cluster $c_{i} \in \mathbb{V}_{x}$ and $c_{j} \in \mathbb{V}_{y}$. The edge between $c_i$ and $c_j$ is denoted as $e_{i,j} :c_i \leftrightarrow  c_j $. There is no edge inside $\mathbb{V}_{x}$ or $\mathbb{V}_{y}$, therefore, our graph $\mathbb{G}$ can be regarded as a bipartite graph as shown in Fig.\ref{rawsegmentation}. 
	
\begin{figure}
		\centering
		\includegraphics[height=2.6cm]{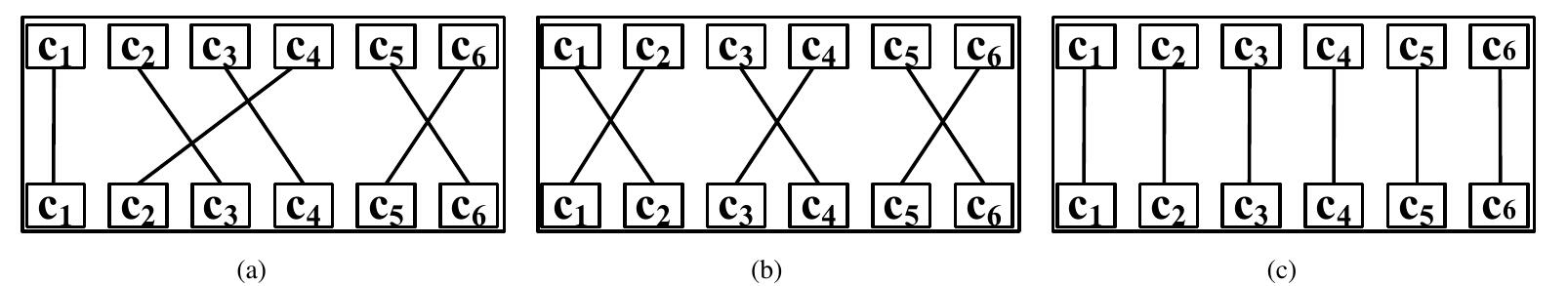}
		\caption{Clustering results indicated by the perfect matching. (a) The combination solution: $\{c_1\},\{c_2,c_3\},\{c_3,c_4\},\{c_4,c_2\},\{c_5,c_6\}$. (b)  The combination solution: $\{c_1,c_2\},\{c_3,c_4\},\{c_5,c_6\}$. (c) The combination solution: $\{c_1\},\{c_2\},\{c_3\},\{c_4\},\{c_5\},\{c_6\}$.}
		\label{matching}
	\end{figure}

	In each hierarchy of the hierarchical clustering, a cluster can be combined with no more than one cluster. In our work, the combination of clusters is indicated by the perfect matching in a bipartite graph. The matching of a graph is a set of edges without common vertices. The perfect matching means that every cluster is connected with a different cluster in the matching. The result of the hierarchical clustering can be determined by a perfect matching as shown in Fig.\ref{matching}.	In our perfect-matching-based representation, the edge $e_{i,j}: c_i \leftrightarrow c_j$, which is between $c_{i} \in \mathbb{V}_x$ and $c_{j} \in \mathbb{V}_y$ under $i \ne j$, means that the cluster $c_i$ and $c_j$ will be combined into one cluster and the edge $e_{i,i}: c_i \leftrightarrow c_i$ means that the cluster $c_i$ is not chosen for the combination in the current hierarchy. 
	
	The above-mentioned Eq.(\ref{cluster}) aims to find the optimal combination which achieves the minimum sum of the dissimilarity in the clustering. In our work, the optimization is achieved by solving the minimum-cost perfect matching in a bipartite graph $\mathbb{G}$. 
	
	Denote the capacity of a perfect matching as $CostPM$, which determines the sum cost of the combination and is calculated as
	
	\begin{align}
	CostPM=\sum_{e_{i,j} \in \Phi} \textbf{PM}\left(c_i, c_j\right), e_{i,j}:c_i \leftrightarrow c_j
	\label{capacity}
	\end{align} 
	
	\noindent where $\Phi$ is the set of edges in the obtained perfect matching. $\textbf{PM}(c_i,c_j)$ aims to calculate the cost of merging clusters that are connected in the obtained matching. Our goal is to find the minimal $CostPM$ for the optimal combination in the hierarchical clustering. $\textbf{PM}(c_i,c_j)$ is obtained from the above-calculated proximity matrix. It is worth noting that $\textbf{PM}(c_i,c_i)$ is weighted by a user-defined cutoff distance $SM$ rather than 0. Details of our optimal hierarchical clustering (OHC) via the minimum-cost perfect matching are shown in Algorithm 1. 
	  	  	
	\begin{algorithm}[t]   
		\caption{Method of the proposed OHC via the minimum-cost Perfect Matching}   
		\label{alg:Framwork}   
		\begin{algorithmic} 	 
			\REQUIRE The point cloud set: $P=\{p_{1},p_{2},p_{3},...,p_{n}\}$. \\                                  
			\ENSURE The cluster set $C=\{c_i,c_2,...,c_t\}$.\\
						
			\STATE Step 1: Initialize $\mathbb{V}_{x}=\{c_{1},c_{2},...,c_{n}\}$ by setting $c_{1}=\{p_{1}\}, c_{2}=\{p_{2}\},..., c_{n}=\{p_{n}\}$;
			
			\textbf{Repeat}
			\STATE Step 2:  Let $\mathbb{V}_{y}=\mathbb{V}_{x}$; 
			
			\STATE Step 3:  Connect the cluster $c_i \in \mathbb{V}_{x}$ with $c_j \in \mathbb{V}_{y}$ to form the edge $e_{i,j}$, where $1 \le i,j \le n$;
			
			\STATE Step 4: Compute the proximity matrix $\textbf{PM}$ using Eq.(\ref{smoothnessterm});\\ 
			
			\STATE Step 5: Optimize Eq.(\ref{capacity}) by solving the minimum-cost perfect matching of $\mathbb{G}=\{\mathbb{V}_{x},\mathbb{V}_{y},\mathbb{E}\}$ using the Kuhn-Munkres algorithm \cite{Munkres1957pp3238};
			
			\STATE Step 6: Combine every connected clusters in $\Phi$ into one cluster;
			
			\STATE Step 7: Update $\mathbb{V}_{x}$ using those combined clusters from Step 6;
			
			\textbf{Until} $\mathbb{V}_{x}$ converges;

			\RETURN $C=\mathbb{V}_{x}$.  
		\end{algorithmic}  
	\end{algorithm} 
	
	In Algorithm 1, steps 1-3 aim to model a bipartite graph $\mathbb{G}=\{\mathbb{V}_x,\mathbb{V}_y,\mathbb{E}\}$. Edges in the initialization are in a full connection, i.e. each $c_i \in \mathbb{V}_{x}$ is connected with all other $c_j \in \mathbb{V}_{y}$. The step 4 is to calculate the dissimilarity between every two clusters to form the proximity matrix.  The step 5 aims to solve the optimization of Eq.(\ref{capacity}). The step 6 is to combine those connected clusters in the obtained perfect matching into one cluster. The step 7 is to update $\mathbb{V}_{x}$ based on the achieved optimal combination solution to move up the hierarchy. 
	
\begin{figure}
		\centering
		\includegraphics[width=16.0cm]{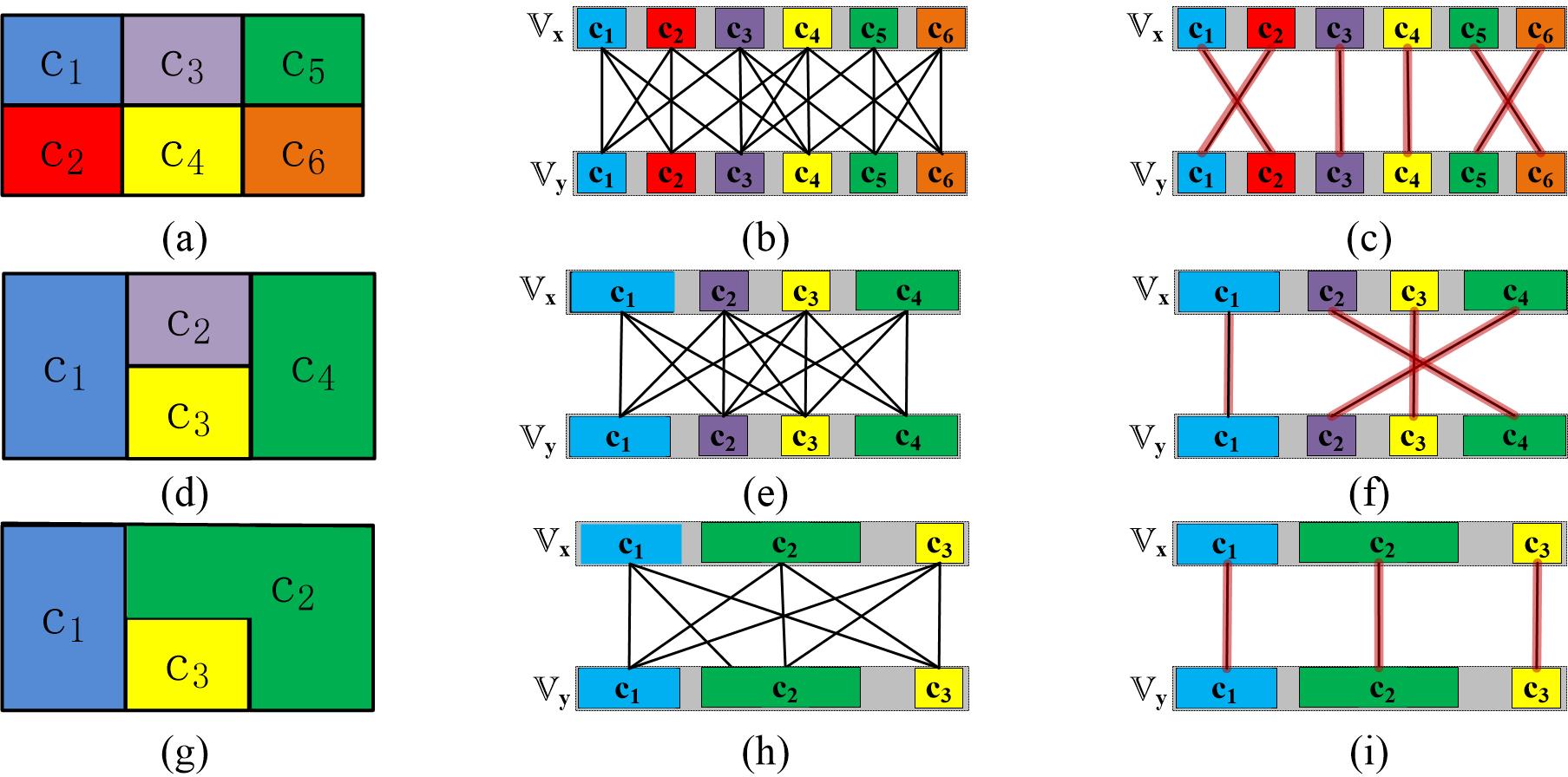}
		\caption{A toy example of the proposed OHC algorithm. (a) The initial clusters. (b) The modeled bipartite graph based on (a). (c) The minimum-cost perfect matching of (b). (d) Clusters after the combination based on (c). (e) The modeled bipartite graph based on (d). (f) The minimum-cost perfect matching of (e). (g) Clusters after the combination based on (f). (h) The modeled bipartite graph based on (g). (i) The minimum-cost perfect matching of (h).}
		\label{optimal}
	\end{figure}

	In the task of the individual object segmentation, spatially non-adjacent clusters are not expected to be combined, thus, connections are only between two adjacent clusters. The following is a toy example of the segmentation based on the proposed Algorithm 1. As shown in Fig.\ref{optimal}(a), the input $P$ is $\{p_{1},p_{2},p_{3},p_{4},p_{5},p_{6}\}$. Step 1: initialize the node set $\mathbb{V}_{x}$ as $\{c_{1},c_{2},c_{3},c_{4},c_{5},c_{6}\}$, where $c_{1}=\{p_{1}\}, c_{2}=\{p_{2}\}, c_{3}=\{p_{3}\}, c_{4}=\{p_{4}\}, c_{5}=\{p_{5}\}$ and $c_{6}=\{p_{6}\}$; Step 2: form the node set $\mathbb{V}_{y}$ the same as $\mathbb{V}_{y}$; Step 3: connect spatially adjacent clusters between $\mathbb{V}_{x}$ and $\mathbb{V}_{y}$ to form the edge set $\mathbb{E}$ as shown in Fig.\ref{rawsegmentation}(b); Step 4: calculate the proximity matrix $\textbf{PM}$ as 
    
    $$\textbf{PM}=\begin{bmatrix}
    	SM & D(\{c_1,c_2\}) & D(\{c_1,c_3) & D(\{c_1,c_4\}) & D(\{c_1,c_5\}) & D(\{c_1,c_6\})\\ 
    	D(\{c_2,c_1\}) & SM & .& . & . & D(\{c_2,c_6\}) \\ 
    	D(\{\{c_3,c_1\})& . & SM & . & . & D(\{c_3,c_6\}) \\ 
    	D(\{c_4,c_1\})& . & . &SM. & . & D(\{c_4,c_6\})\\ 
    	D(\{c_5,c_1\}) & . & . & . & SM & D(\{c_5,c_6\})\\ 
    	D(\{c_6,c_1\}) & D(\{c_6,c_2\}) & D(\{c_6,c_3\}) & D(\{c_6,c_4\}) & D(\{c_6,c_5\}) & SM
    \end{bmatrix}_{6 \times 6}$$
     
\begin{figure}
			\centering
			\includegraphics[width=16cm]{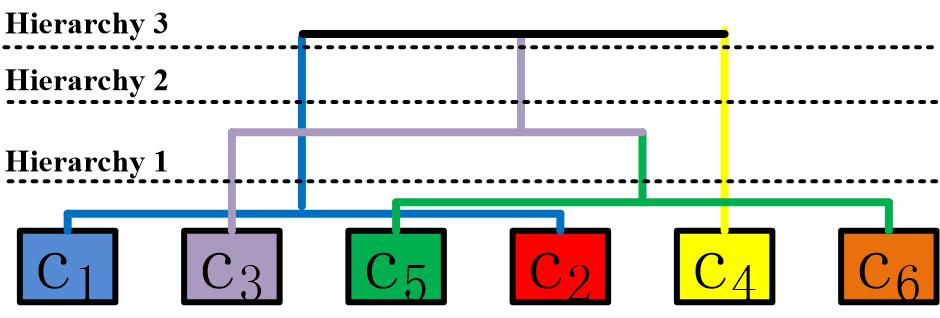}
			\caption{The dendrogram of the example clustering.}
			\label{hierarchy}
		\end{figure}

    \noindent Step 5: solve the minimum-cost perfect matching of $\mathbb{G}$ as shown in Fig.\ref{optimal}(c). Edges in the obtained perfect matching is $\Phi=\{e_{1,2},e_{2,1},e_{3,3},e_{4,4},e_{5,6},e_{6,5}\}$; Step 6: combine clusters $c_1 \in \mathbb{V}_{x}$ and $c_2 \in \mathbb{V}_{y}$ into $c_1$, and $c_5 \in \mathbb{V}_{x}$ and $c_6 \in \mathbb{V}_{y}$ into $c_4$. The cluster $c_3$ and $c_4$ are unchanged and renamed as $c_2$ and $c_3$, respectively. Step 7: update $\mathbb{V}_{x}$ as $\{c_1,c_2,c_3,c_4\}$; Now the first iteration is done. Repeat Step 2-7 and finally $\mathbb{V}_{x}$ converges to $\{c_1,c_2,c_3\}$ as shown in Fig.\ref{optimal}(i); Step 8: return $\mathbb{V}_{x}$ as the clustering $C$. The dendrogram of the above hierarchical clustering is shown in Fig.\ref{hierarchy}

	\section{Experimental results}
	
	\subsection{Evaluation methods}
	
	Suppose that the cluster result is $C=\{c_{1}, c_{2},..., c_{i},..., c_{t}\}$ and ground truth is $C'=\{c'_{1}, c'_{2},..., c'_{j},..., c'_{t'}\}$. Each $c_{i}$ or $c'_{j}$ means a cluster of points and there are $t$ clusters in $C$ and $t'$ clusters in $C'$. For the purpose of evaluating our segmentation results, we define the completeness $n_{com}$ and correctness $n_{cor}$ as Eq.(\ref{PreRec}). 
	
	\begin{align}
	n_{com}=\frac{1}{t} \sum_{i=1}^{t} (\frac{ \max \limits_{j=1}^{t'} |c_{i} \bigcap c'_{j}|}{|c_{i}|}) \notag \\
	n_{cor} =\frac{1}{t'} \sum_{j=1}^{t'} (\frac{ \max \limits_{i=1}^{t} |c'_{j} \bigcap c_{i}|}{|c'_{j}|})
	\label{PreRec}
	\end{align}
	\noindent There are two steps in the calculation of $n_{com}$. The first step is to achieve the maximum of $|c_i \cap c'_j|/|c_i|$ where $i$ is a constant and $j$ is from 1 to $t'$. This is the completeness of each cluster in $C$. The second step is to obtain the mean of values from the first step. Similar steps are used to calculate $n_{cor}$. The completeness $n_{com}$ is to measure the ratio between the correctly grouped points and the total points in the result. The correctness $n_{cor}$ is to measure the ratio between the correctly grouped points and the total points in the ground truth. Both the completeness and correctness range from 0 to 1. 
	
	Our completeness and correctness are similar to the purity index \cite{manning1995introduction} which is an evaluation measure by calculating the similar between two clusters. The difference is that we focus on the ratio of correctly grouped points and total points which is a commonly used criterion in the segmentation evaluation as shown in \cite{YangFangLi2013pp8093} \cite{BoykoFunkhouser2011pp212} \cite{KumarMcElhinneyLewisEtAl2013pp4455}. The problem of Eq.(\ref{PreRec}) is that if there is only one cluster in the ground truth $C'$, $n_{com}$ will be always 1, and if there is only one cluster in the result $C$, $n_{cor}$ will be always 1. In order to address this problem, we choose the minimum of $n_{com}$ and $n_{cor}$ as the segmentation accuracy, i.e. $n_{acc}=\min(n_{com}, n_{cor})$, to measure the difference of points between $C$ and $C'$. To combine the completeness and correctness, the criterion $F_1$-score, i.e. $n_{F1}=2 \times ( n_{cor}  \cdot n_{com})/( n_{cor} + n_{com})$, is used to favor algorithms with a higher sensitivity and challenges those with a higher specificity \cite{GoutteGaussier2005pp345359}. 
	
	\begin{figure}
		\centering
		\includegraphics[width=10.0cm]{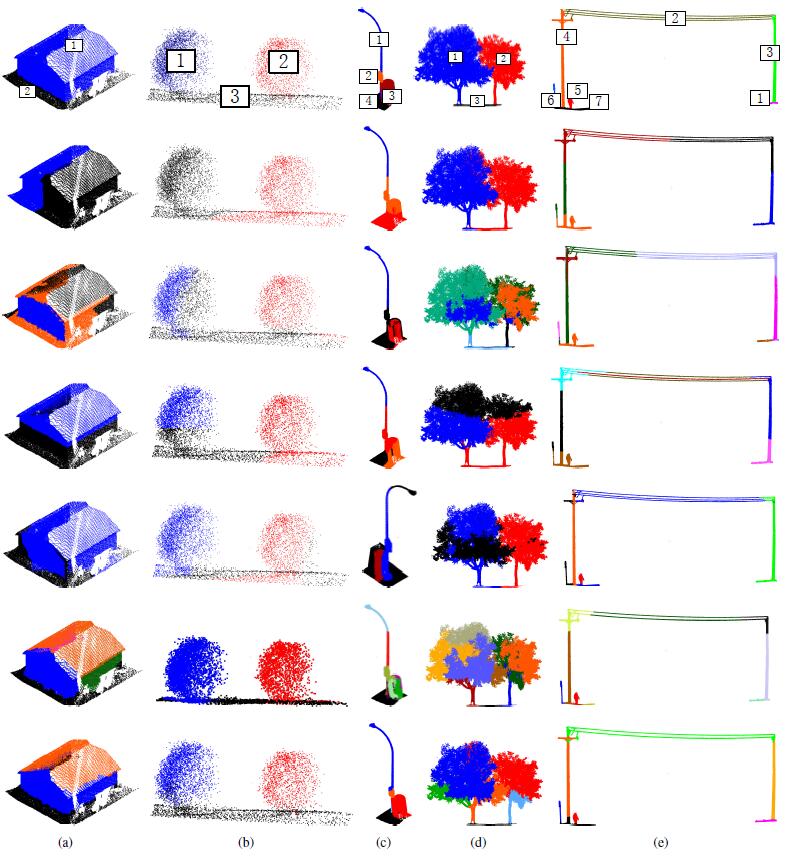}		
		\caption{Performances of different segmentation methods. (a) HouseSet. (b) BushSet. (c) LamppostSet. (d) TreeSet. (e) PowerlinesSet. Row 1:  the segmentation ground truth of each scene. Row 2:  performance of KMiPC\cite{isprs-archives-XLII-1-W1-595-2017}. Row 3:  performance of KNNiPC\cite{douillard2011segmentation}. Row 4:  performance of 3DNCut\cite{YuLiGuanEtAl2015pp13741386}. Row 5:  performance of MinCut\cite{GolovinskiyKimFunkhouser2009pp21542161}. Row 6:  performance of PEAC\cite{feng2014fast}. Row 7:  performance of the proposed OHC.}
		\label{comparsions}
	\end{figure}

	\subsection{Comparisons of methods}
	This section will evaluate performances of KMiPC\cite{isprs-archives-XLII-1-W1-595-2017}, KNNiPC\cite{douillard2011segmentation}, 3DNCut\cite{YuLiGuanEtAl2015pp13741386}, MinCut\cite{GolovinskiyKimFunkhouser2009pp21542161}, PEAC \cite{feng2014fast}  and the proposed OHC on five typical scenes to show our superiority. The selected scenes for testing are shown in Fig.\ref{comparsions}, including the HouseSet (2 labels): a single object, the BushesSet (3 labels): two separated sparse objects, the LamppostSet (4 labels): two connected rigid objects, the TreesSet (3 labels): two connected non-rigid objects and the PowerlinesSet (7 labels): a complex scene with different objects. 
	
	The first row of Fig.\ref{comparsions} shows the segmentation ground truth of each scene. The ground truth is obtained by manually segmenting visual independent objects. From the second row to the last row are the performance of KMiPC, KNNiPC, 3DNCut, MinCut, PEAC and the proposed OHC, respectively. In the visualization, we use different colors to represent distinct segments. KMiPC, KNNiPC and MinCut are implemented by the Point Cloud Library (www.pointclouds.org/), 3DNCut is extended from the normalized-cut method (www.cis.upenn.edu/~jshi/software/). PEAC is achieved based on the software of Feng et al. (www.merl.com/research/?research=license-request\&sw=PEAC).
	
	KMiPC is suitable for segmenting symmetric objects and works well in the BushesSet and TreesSet. However, it fails to split connected objects. KNNiPC relies on the density heavily and causes that an object is clustered into different groups as shown in the TreesSet. 3DNCut tries to normalize the difference of points' Euclidean distances in each group. Therefore, it tends to group points evenly as shown in the HouseSet and TreesSet. MinCut performances well in most cases when the required center point and radius of each object are set properly. PEAC clusters points based on the plane information and it asks for a region growing process to obtain an individual object. 
	
\begin{figure}
			\centering
			\includegraphics[width=10cm]{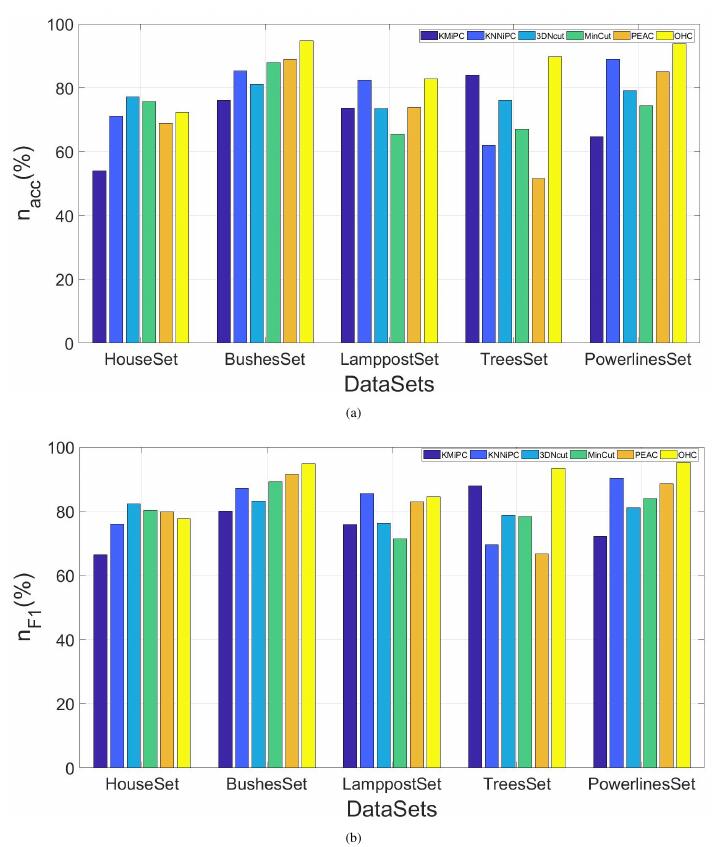}
			\caption{Evaluation of different methods. (a) Accuracy $n_{acc}$. (b) $F_1$-score $n_{F1}$.}
			\label{evaluation}
		\end{figure}

	The proposed OHC is not sensitive to the density of points as shown in the BushesSet. Moreover, it does not require for presetting the number of clusters and the location of each object to achieve the multi-object segmentation. The attached traffic sign in the LamppostSet is segmented successfully using the proposed OHC. The segmentation of the overlapping region between two non-rigid objects is rather difficult due to the rapid change of normal vectors as shown in the TreesSet. The quantitative evaluation is shown in Fig.\ref{evaluation}. One can observe that our OHC is more accurate than other methods in most experiment scenes. 
	
		\begin{figure}
			\centering
			\includegraphics[width=13cm]{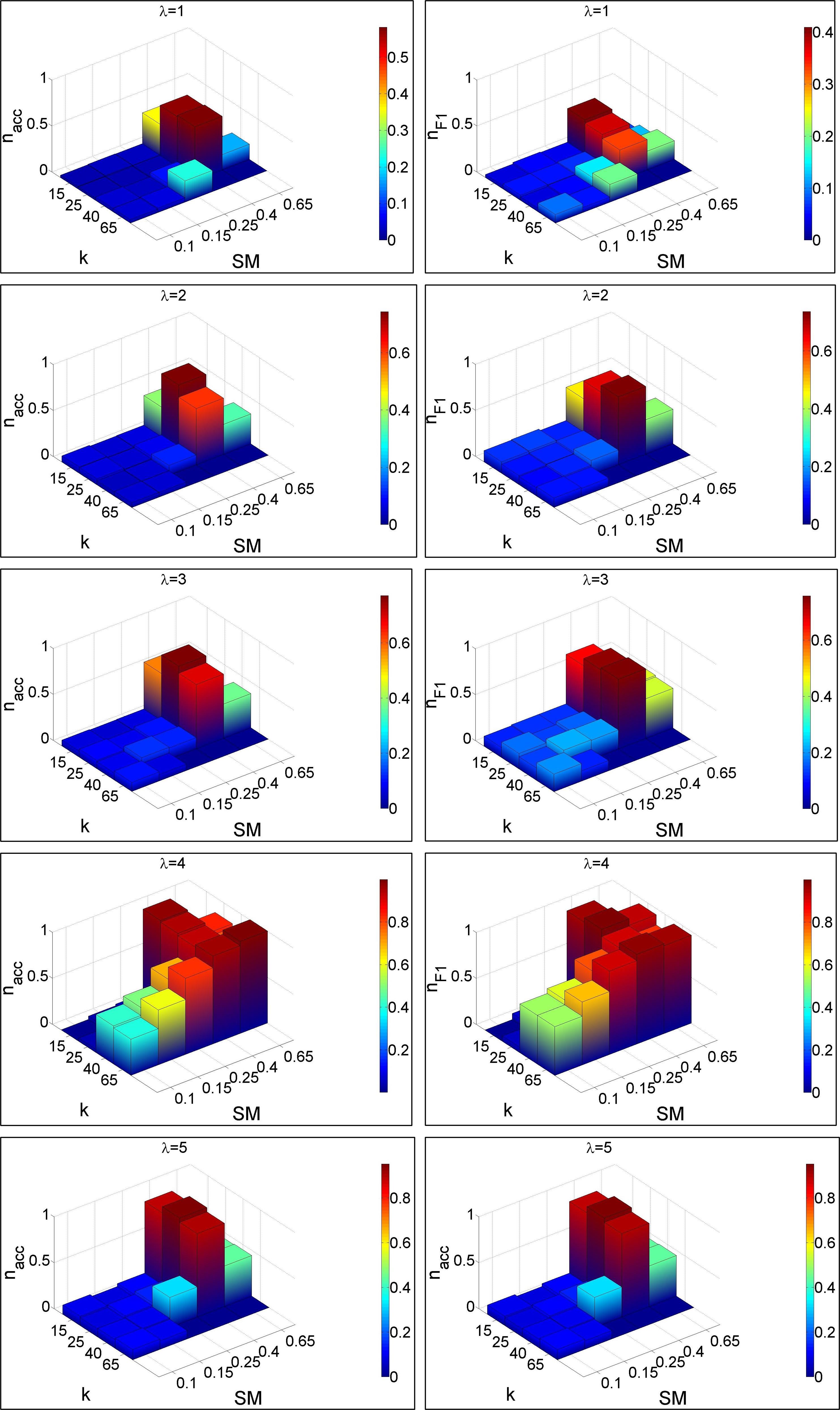}	
			\caption{Parameters setting. $k$ is chosen from \{15, 25, 40, 65\} and $SM$ is chosen from \{0.1, 0.15, 0.25, 0.4, 0.65\}. (a) $\lambda=1$. (b) $\lambda=2$. (c) $\lambda=3$. (d) $\lambda=4$. (e) $\lambda=5$.}
			\label{paraset}
		\end{figure}
	 
	In the implementation of OHC, there are three parameters, namely the $k$ to select the nearest neighbor points, $\lambda$ to balance the weight of the $\alpha\left(p_i,p_j,c_i,c_j\right)$ and $\beta\left(p_i,p_j,c_i,c_j\right)$, and $SM$ for calculating the proximity matrix. A large $k$ or $SM$ may cause the appearance of the under-segmentation and a small $k$ or $SM$ will increase the over-segmentation rate in results. A large $\lambda$ works well for segmenting connected objects and a small $\lambda$ is preferred when there are less overlapping objects. Evaluation experiments of parameters are shown in Fig.\ref{paraset}. In our work, $\lambda$ is 4, $k$ is 40 and $SM$ is 0.4.
	
\begin{figure}
		\centering
		\includegraphics[width=16cm]{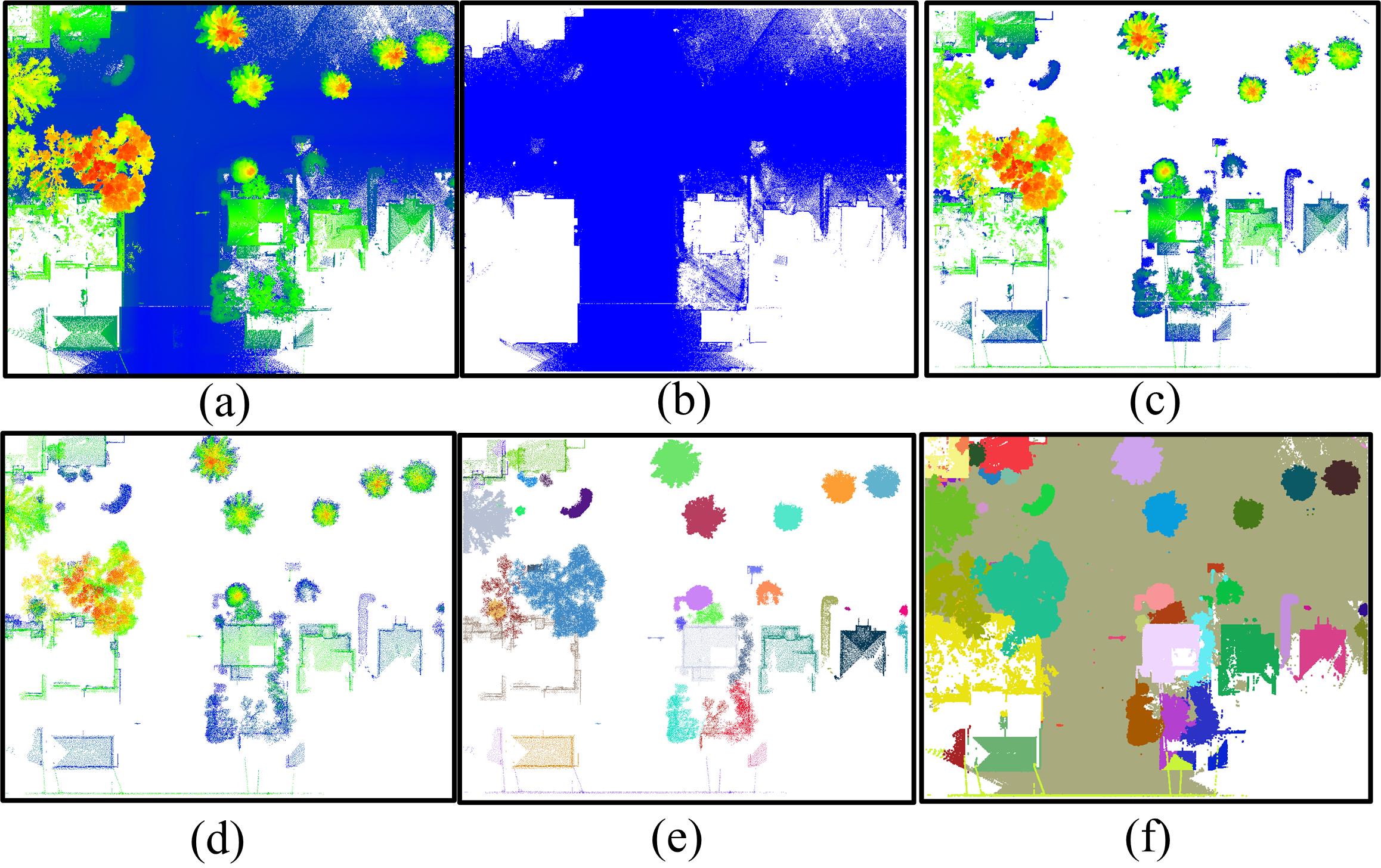}
		\caption{An example of segmenting a large-scale point set. (a) Input scene. (b) Ground points. (c) Off-ground points. (d) Down-sampling data (10\% points of (c)). (e) Performance of the proposed OHC on (d). (f) Segmentation results of (a).}
		\label{framework}
	\end{figure}

\begin{figure}
		\centering
		\includegraphics[height=10cm]{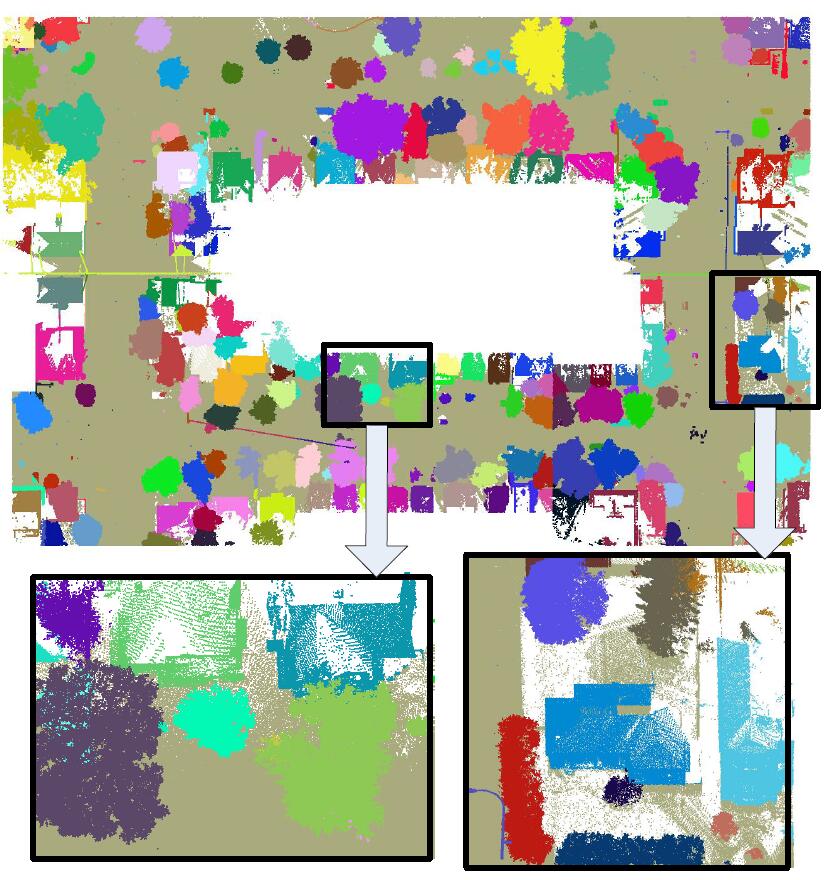}
		\caption{Results of a large-scale residential point set.}
		\label{largescale1}
	\end{figure}

\begin{figure}
		\centering
		\includegraphics[height=10cm]{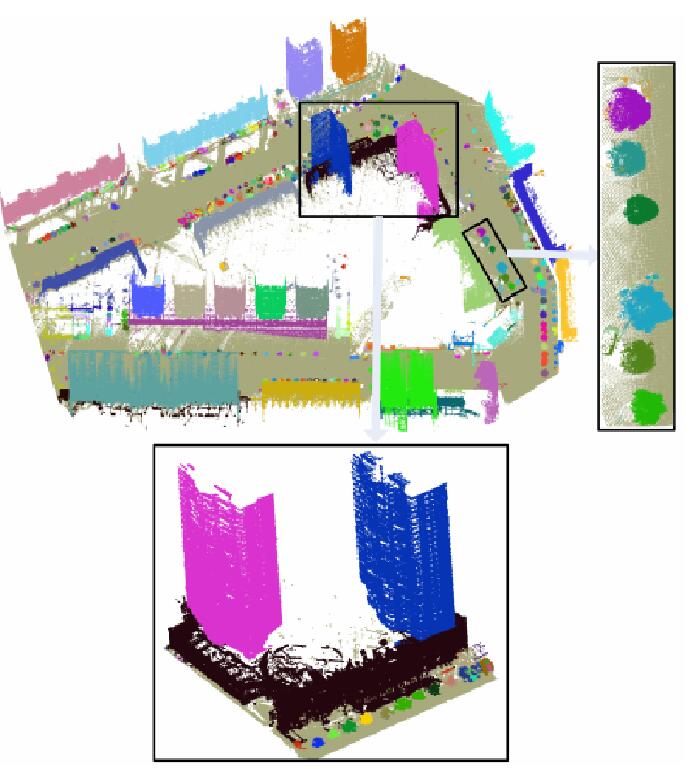}
		\caption{Results of a large-scale urban point set.}
		\label{largescale2}
	\end{figure}

	For dealing with a large-scale scene, we present a framework to increase the efficiency of OHC: 1) remove ground points using the Cloth Simulation Filter (CSF) approach \cite{zhang2016easy}; 2) down-sample the off-ground points into a sparse point set; 3) apply the proposed OHC on the above down-sampling data; 4) assign those unlabeled points in the original off-ground point set with the same label as their nearest labeled point. An example is described in Fig.\ref{framework}. Fig.\ref{framework} (a) shows the input point clouds.  Fig.\ref{framework} (b) and (c) are filtered ground and off-ground points, respectively.  Fig.\ref{framework} (d) is the down-sampling points of Fig.\ref{framework}(c). Fig.\ref{framework} (e) is the performance of OHC on the down-sampling points and Fig.\ref{framework} (f) is the final results of the input scene. Fig.\ref{largescale1} shows the performance of the proposed OHC on a large-scale residential point set (558 MB, 12,551,837 points) which is obtained by the RIEGL scanner. Fig.\ref{largescale2} shows our results of a large-scale urban point set (528 MB, 10,870,886 points) collected by the Optech scanner.

	\subsection{Discussion of the proposed OHC}
	
	The proposed OHC algorithm works automatically in the multi-object segmentation. We do not need to preset any parameters manually, e.g. the initial number of clusters or the location of each object. Our performance is competitive against the state-of-the-art methods. The proposed optimal hierarchical clustering is general in the segmentation of point clouds but can be a disadvantage in terms of speed. The complexity of OHC relies on the Kuhn-Munkres algorithm which is $O(N^{3})$. The above-mentioned experiments were done on a Windows 10 Home 64-bit, Intel Core i7-4790 3.6GHz processor with 16 GB of RAM and computations were carried on Matlab R2017a. It took us around four hours to achieve the segmentation of the large-scale residential point set (12,551,837 points) and the urban point set (10,870,886 points).
	
	\section{Conclusions}

	This paper investigates a new optimal hierarchical clustering method (OHC) for segmenting multiple individual objects from 3D mobile LiDAR point clouds. The combination of clusters is represented by the matching of a graph. The optimal combination solution is obtained by solving the minimum-cost perfect matching of a bipartite graph.	The proposed algorithm succeeds to achieve the multi-object segmentation automatically. We test our OHC on both a residential point set and an urban point set. Experiments show that the proposed method is effective in different scenes and superior to the state-of-the-art methods in terms of the accuracy.
	
	Future work will focus on adding the dissimilarity measure of the intensity and color information of points in the calculation of the proximity matrix. Besides, we will try to improve the efficiency of the combination by using the supervoxel technique in the clustering.

\bibliographystyle{IEEEtran} 
\bibliography{Myreview}

\end{document}